# Theatre in the Loop: A Rehearsal-Based, Collaborative Workflow for Expressive Robotic Behaviours

Pavlos Panagiotidis[1] [0009-0000-8229-7239], Victor Zhi Heung Ngo[1] [0009-0003-0805-5292], Sean Myatt[2], Roma Patel[3] [0000-0003-2075-2560], Rachel Ramchurn[3] [0009-0005-3271-4467], Alan Chamberlain[1] [0000-0002-2122-8077], Ayse Kucukyilmaz[1] [0000-0003-3202-6750]

[1] School of Computer Science, University of Nottingham, Nottingham, United Kingdom
`{victor.ngo, pavlos.panagiotidis, alan.chamberlain, ayse.kucukyilmaz}@nottingham.ac.uk`
[2] School of Art & Design, Nottingham Trent University, Nottingham, United Kingdom
`sean.myatt@ntu.ac.uk`
[3] Makers of Imaginary Worlds, Nottingham, United Kingdom
`{roma, rachel}@makersofimaginaryworlds.co.uk`

**Abstract.** In this paper, we propose *theatre-in-the-loop*, a framework for developing expressive robot behaviours tailored to artistic performance through a director-guided puppeteering workflow. Leveraging theatrical methods, we use narrative objectives to direct a puppeteer in generating improvised robotic gestures that convey specific emotions. These improvisations are captured and curated to build a dataset of reusable movement templates for standalone playback in future autonomous performances. Initial trials demonstrate the feasibility of this approach, illustrating how the workflow enables precise sculpting of robotic gestures into coherent emotional arcs while revealing challenges posed by the robot's mechanical constraints. We argue that this practice-led framework provides a model for interdisciplinary teams creating socially expressive robot behaviours, contributing to (1) theatre as an interactive training ground for human-robot interaction and (2) co-creation methodologies between humans and machines.

## 1 Introduction

This paper presents a method developed through the creative process of designing NED II, an evolution of the Never-Ending Dancer (NED), a robotic performer originally created for The Thingamabobas art installation [1]. The original NED (Fig. 1, left), a costumed robot featured in the installation (Fig. 1, right), successfully attracted and engaged child audiences through its playful design and choreographed motion. However, in-the-wild studies with 18 children [2] demonstrated several critical limitations: NED lacked expressive nuance, failed to adapt responsively to user input, and often disengaged audiences through repetitive or contextually ambiguous gestures. For example, children frequently grew confused or frustrated when NED entered its default "sleep" state, or when it failed to mirror playful gestures like dancing or waving. One child notably remarked, *"Why does it keep going to sleep?"*, a reflection of both unmet expectations and a perceived lack of reciprocity. These moments revealed that although NED was compelling as a kinetic sculpture, it struggled to perform with narrative coherence or emotional depth.

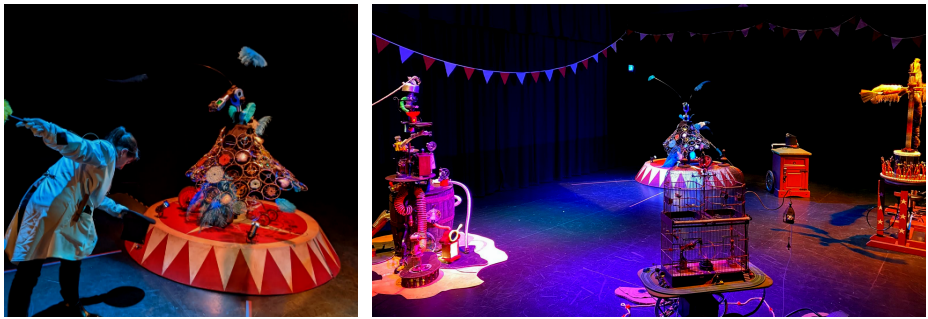

**Fig 1.** (left) The Never-Ending Dancer (NED) positioned on a theatrical stage, interacting with a performance artist. (right) The interactive art installation Thingamabobas.

These shortcomings highlight broader challenges in designing robots for performative interaction: how can a robot convincingly convey intentionality, sustain engagement through expressive movement, and adapt to live dramaturgical contexts? While many approaches in Human–Robot Interaction (HRI) focus on static behaviour scripting or emotion recognition, they often omit the embodied, improvisational, and context-sensitive practices that define performance-making.

To address this gap, we introduce *theatre-in-the-loop*, a rehearsal-driven method for developing expressive robotic motion grounded in collaborative theatrical practice. This approach integrates directorial framing, real-time puppeteering, and iterative refinement to shape emotionally resonant gestures that are both situated and reusable. The resulting motion library offers a flexible foundation for standalone expressive behaviour in future performance contexts.

## 2     Related Work: Expressive Robotics and Theatrical Methods

Recent studies have employed data-driven and language-based frameworks to explore the generation of expressive robot behaviour. Panteris et al. [3] proposed a probabilistic model for generating varied wave gestures on a 6-degree-of-freedom (DOF) arm. This model demonstrated the potential of incorporating stochasticity to introduce natural variation to repetitive motions. Zhou et al. [4] investigated the expressive potential of temporal modulation, including variations in speed and deliberate pauses, in conveying internal state transparency. Building upon these concepts, Huang et al. [5] proposed the EMOTION system, which employs large language models (LLMs) to sequence contextually appropriate gestures, thereby integrating verbal cues with non-verbal expression. In their seminal study, Tang and Dondrup [6] examined the process of tri-modal gesture generation by integrating speech, vision, and semantic information. Building on this research, Schreiter et al. [7] implemented multimodal intention communication on industrial manipulators with the aim of improving human interpretability. Mahadevan et al. [8] further demonstrated that natural-language prompts can be translated into parameterised motion sequences, thereby enabling generative, adaptive behaviours in social contexts.

Although these approaches advance the technical foundations for expressive motion generation, they often lack direct creative input from practitioners with expertise in performance, embodiment, and narrative. As a result, the generated behaviours, though technically varied, can fall short in terms of intentionality, emotional richness, and contextual meaning. Some studies have addressed this by using live performance as both inspiration and testbed for socially expressive robotics. For example, Katevas et al. [9] analysed stand-up comedy and incorporated feedback from comedians to pre-script a robot's gaze and gestures, staged these behaviours in a live performance, and measured audience responses. In contrast, our *theatre-in-the-loop* approach engages the early stages of the creative process, drawing on performance expertise to shape robotic expression while embedding interdisciplinary collaboration directly into the generative workflow. By integrating embodied improvisation, narrative framing, and iterative evaluation within a rehearsal setting, it fosters behaviours that are not only expressive but also creatively authored and contextually resonant within a performative context.

## 2.1 Theatre Techniques in HRI

Theatre methods have increasingly been used to conceptualise and structure robotic behaviours. Hoffman [10] draws parallels between established acting methodologies and the design of socially intelligent robots, identifying four critical acting principles he considers applicable to human-robot interaction (HRI): *Psycho-physical unity* challenges conventional distinctions between cognition and motion in robotics, promoting an embodied intelligence model where gestures actively shape cognitive processes. *Mutual responsiveness* inspired by Meisner's relational performance approach [11], suggests robots should dynamically adapt their actions in real-time to ongoing interactions. *Objectives and inner monologue* advocate for robotic behaviours driven by implicit goals and reflective states, enhancing narrative coherence and intentionality. Finally, *context and given circumstances* underscore the importance of designing adaptive robotic behaviours, responsive to evolving environmental conditions, ensuring contextually appropriate expressions.

Beyond conceptual framing, modular theatre methods have also been employed by designers to fine-tune expressive qualities, such as posture, breath, and tempo. Laban Movement Analysis (LMA) for instance, provides a structured vocabulary to break down gesture into components like weight, flow, and time [12]. Building on LMA, Laviers and Maguire [13] introduce BESST, a system for documenting expressive

4      Pavlos Panagiotidis, Victor Zhi Heung Ngo, Sean Myatt, Roma Patel, Rachel Ramchurn, Alan Chamberlain, and Ayse Kucukyilmaz

movement through a notation framework that emphasises embodied observation and structured analysis, providing a method to encode and analyse expressive behaviours in HRI.

Such theatre-inspired approaches form a performative framework for HRI; one that positions the robot as a dramaturgically scaffolded agent. Our work shapes a workflow through an interdisciplinary collaboration between a theatre director, a puppeteer, roboticists, and designers responsible for the design of the performative installation NED II. By adapting theatre rehearsal methods to facilitate improvisational puppeteering during rehearsal, we aim to endow the robot with human-driven responsiveness and performative presence.

### 2.2    Puppetry and Directing as a Model for Enhancing Robotic Expressivity

Key challenges in robotic behaviour design for performance include the difficulty of conveying intentionality, emotional nuance, and narrative coherence. We propose that a) puppetry, which excels at animating inanimate objects through real-time, tangible manipulation, and b) theatre directing, which shapes narrative structure, emotional arcs, and performance clarity by guiding the rehearsal process, together offer powerful models to address these challenges.

Rather than striving for anthropomorphic realism, puppeteers can convey intentionality and expressivity through rhythm, gesture, and stylised movements. What makes this approach particularly appealing in HRI is its compatibility with learning from demonstration frameworks, which can be trained with improvisational data to generate autonomous robot behaviours learned from puppeteer actions [14]. Many existing studies focus on expressive movement but rely on pre-authored behaviours, overlooking live, collaborative or dramaturgically driven improvisation. For example, a hybrid puppetry-based control system for robotic stage actors [15] integrated pre-scripted animations, parametric behaviours, and real-time eye-contact inverse kinematics to enable expressive interactions with human performers. As such, puppetry's potential for embodied, improvisational exploration, remains largely untapped in current robotics frameworks.

Similarly, the deeper dramaturgical potential of theatre directing has remained underexplored in HRI, with directors typically involved in peripheral or project-specific capacities rather than integrated deep into the creative process. Hoffman et al. [15] collaborated with a theatre director to refine gesture animations created in 3D software, structuring scenes into beats with precisely timed joint movements. Similarly, Zeglin et al. [16] involved a director to help refine animator-generated gestures, so they could be effectively translated to a mobile manipulator with two anthropomorphic arms. While these studies demonstrated enhanced expressive quality, they limited the director's contribution to refining pre-existing material rather than involving them directly in a generative, improvisational creation of robotic behaviour. In contrast, our approach integrates the director's dramaturgical expertise from the outset, shaping robotic gestures through rehearsal practices.

Related work by Rozendaal et al. [17] also explores the incorporation of theatrical expertise in HRI by introducing a mixed-reality stage, where theatre professionals co-design behaviours through embodied interaction in virtual reality. Their approach reframes robotic limitations as creative constraints, enabling situated, repeatable design processes that integrate narrative intent and expressive motion within a transdisciplinary, performative framework. Building on this, we incorporate real-time improvisation and dramaturgical collaboration directly into the physical rehearsal process.

### 2.3    Gaps in the Domain

Despite growing interest in theatre-inspired methods for refining or prototyping expressive robotic gestures, existing approaches remain largely theoretical, fragmented, or confined to virtual spaces. Technical frameworks often lack direct creative input from performance practitioners, while theatre-informed systems tend to retrofit directors' and puppeteers' expertise rather than embedding it from the start. As a result, three key gaps persist in translating theatrical intent into cohesive, reusable robotic behaviours:

1. **Intent-to-Specification Translation:** There are no widely adopted method to convert theatre directing strategies into robot-executable behaviours to support expressive, context-sensitive robotic performance.
2. **Real-time Co-Creation:** Real-time co-creation of expressive, performative robotic behaviour remains challenging due to the absence of a systematic workflow that integrates the expertise of theatre-makers and designers throughout the creative process.
3. **Reuse of Improvisation:** Improvised gestures are often ephemeral, lacking mechanisms for capturing their reuse in future contexts.

## 3    Methodology: A Director, Puppeteer, Roboticist Workflow for Expressive Motion Capture

To address the identified gaps, we propose a creative workflow that adopts a practice-led, interdisciplinary methodology, aimed at generating and capturing expressive robotic behaviour. Drawing from a performative-materialist perspective [18] and mixed-reality speculative design [17], *theatre-in-the-loop* workflow integrates theatrical rehearsal-based embodied improvisation and collaborative performance-making.

The workflow was implemented across two exploratory workshops and developed through collaborative efforts of an interdisciplinary team with established expertise in the domains of performance and technology. The team consists of roboticists specialising in Human-Robot Interaction within artistic settings, a puppeteer with training in design and digital media, a theatre director with research experience in HCI and dramaturgy, and designers with extensive experience in interactive performance technologies. Fig 2 shows an overview of the proposed workflow.

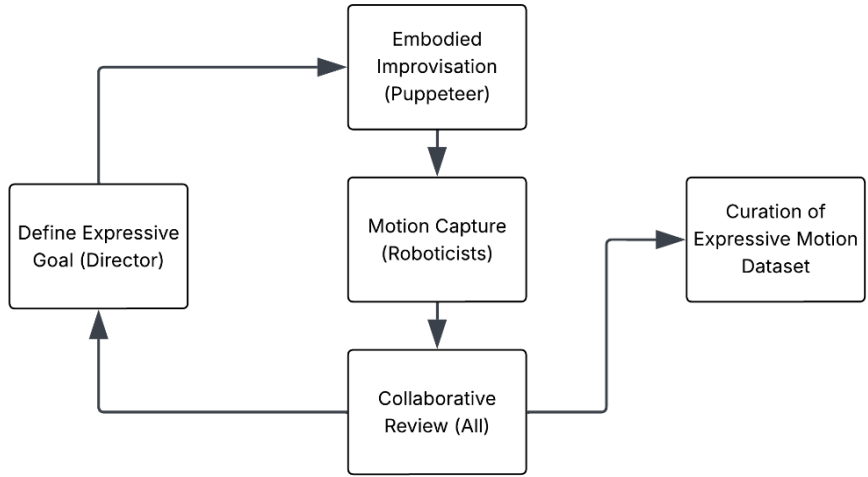

**Fig 2.** An overview of the *theatre-in-the-loop* workflow for expressive robot motion design.

### 3.1 Roles and Contributions

During the implementation of our proposed workflow, the roboticist led the overall design and facilitation of the workshop, shaping both its technical infrastructure and its integration with creative practice. This included configuring the robotic platform, ensuring system safety and responsiveness, and aligning the robot's capabilities with the rehearsal process. The workshop directly supported the roboticist's research aims, centring their investigation into how expressive, reusable robotic behaviours can be generated through live, interdisciplinary collaboration.

In this context, the director translated expressive goals into directing prompts that iteratively shaped improvisations into distinct narrative arcs. The puppeteer interpreted these prompts as motivations for action, physically manipulating the robot to produce expressive nuances (tremors, pauses, and hesitations) captured as reusable motion data. The designers contributed dramaturgical framing, scenographic insights, and creative-critical feedback on gesture expressiveness. The recorded movements were replayed by the robot and iteratively reviewed and refined by the team to enhance narrative coherence and emotional fidelity. Finally, the refined motion data was annotated and stored, forming a curated library of expressive templates for future reuse, including as training material for generative models.

### 3.2 Robotic System and Data Capture

**Robotic Platforms:** In Workshop 1, we used a 7-degree-of-freedom (DOF) Franka Emika Panda arm for physical familiarisation. In Workshop 2, a 6-DOF Universal Robots UR3e collaborative robotic arm was used for gesture creation and data capture. The Franka arm was utilised to facilitate the introduction of the puppeteer to robotic arms, as the UR3e supporting software was not operational at the time of the initial workshop. The UR3e was utilised in the subsequent workshop, as it is the robotic platform that has been selected by the designers for the development of the NED II project.

**Control Interface:** In both workshops, the operation of the robots was achieved through direct physical manipulation. The Franka arm was operated in gravity-compensation mode, while the UR3e was guided via *Freedrive* mode, enabling the puppeteer to manipulate the arm with ease and precision. This configuration enabled the puppeteer to focus on expressive gesture without the need for abstracted control inputs.

**Trajectory Recording and Playback Tool:** Motion trajectories from the Franka arm were recorded using the Franka Control Interface[1] (FCI) through a custom Python script for recording joint position data. Recorded trajectories were not played back during the first workshop.

Recording of the motion trajectories from the UR3e were captured in real-time using *Freedrive* mode, activated using the Teach Pendant. The trajectory recording and playback tool consists of three scripts written in Python using ROS 2[2]. The script (based on *DynamicTrajectoryExecutor* class [19]) logged raw time-stamped joint positions to a CSV file. A second script implemented a timestamp normalisation step to convert absolute timestamps into relative *time_from_start* values required for trajectory execution. No smoothing or trajectory optimisation was applied, preserving the original motion dynamics. Finally, a third script loaded the trajectory and replayed it via ROS 2's *FollowJointTrajectory* action interface, sending commands to the */scale_joint_trajectory_controller/follow_joint_trajectory* topic. This allowed for direct reproduction of the recorded motions without modification, ensuring fidelity to the original performance. The system was developed using Ubuntu (22.04) and ROS 2 (Humble).

## 4     Workshop Structure and Findings

The proposed workflow emerged and was tested across two structured workshops. As this was an exploratory study using a novel method, we selected Ekman's six basic emotions [20] to provide a simple, widely recognised emotional vocabulary that served both as a conceptual foundation and a practical guide for the creative team, enabling consistent mapping of emotional intent to robotic responses.

### 4.1     Workshop 1: Collaborative Foundations for Expressive Robotics

This initial workshop (W1) served as an open exploration of how a director, puppeteer, roboticists and designers can co-create expressive robotic motion. The session focused on exploring the interplay between theatrical improvisation, physical manipulation, and robotic constraints, identifying what creative practices held promise for deeper study, shown in Fig 3 (left).

---

[1]   The Franka Control Interface (FCI) is an API that provides low-level, real-time access to the Franka Emika robot's sensors and actuators for precise motion and force control. https://frankaemika.github.io/docs/

[2]   ROS 2 is an open-source framework for building modular, real-time, and distributed robot applications. https://www.ros.org/

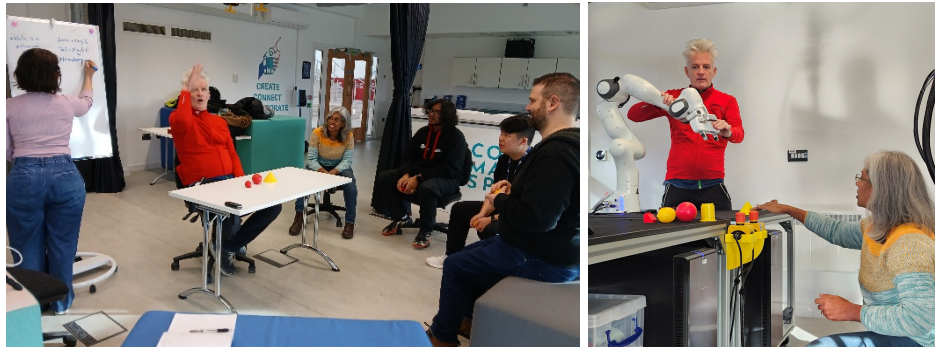

**Fig 3.** (left) The puppeteer, director, roboticists, and designers discuss and engage practically in embodied expressive motion. (right) The puppeteer physically manipulates the Franka Emika Panda robotic arm to engage expressively with the designer.

Using a Franka Emika Panda arm in gravity-compensation mode, activated through the FCI, the puppeteer familiarised himself with the robot and developed an intuitive sense of its motion range and resistance. This initial hands-on engagement proved critical; the puppeteer's physical dialogue with the machine revealed nuances regarding the potential for expressive manipulation that theoretical discussion alone could not. He characterised his interaction with the robot arm as a continuous process of adjustment, watching its motion, sensing its weight and position, and modifying gestures.

The director introduced narrative prompts to shape the interaction, continuously observing, interpreting, and adjusting improvisations in real-time rather than following preplanned scenarios. Simple objects, a ball and a cup, (Fig 3 right), were introduced by the designers and became emotional anchors through framing: a neutral sphere might transform into "disgusting medicine" or a "precious heirloom," compelling the puppeteer to reinterpret gestures in real-time. These improvisations avoided static emotional labels (e.g., "show anger") in favour of dynamic scenarios, such as reluctance giving way to curiosity, or conflict resolving into uneasy acceptance. The team noted how this narrative spontaneity produced gestures that felt more fluid and intentionally grounded than predefined movement primitives. The workshop surfaced two key insights:

**Collaborative Workflows:** How role dynamics (director: narrative, puppeteer: physical interpreter, roboticists/designers: constraint mediators) form a creative loop.

**Improvisation vs. Preset Emotion:** Spontaneous, context-dependent gestures (e.g., "disgust" tied to a specific narrative) felt more authentic than pre-scripted emotional motion primitives, basic units of expressive movement associated with particular emotional states. This session raised the questions: How can we retain the authenticity of improvised gestures in autonomous reproduction? What dramaturgical structures might bridge live, directed puppeteering and programmed behaviour?

### 4.2   Workshop 2: Co-shaping Expressive Robotic Motion

Building on the somatic insights gained in the first session, in Workshop 2 (W2), we used the UR3e robot arm to generate gestures for each of six basic emotions defined by Ekman: anger, disgust, fear, happiness, sadness and surprise.

**W2 Session 1 - Human Expression Study:** This explored how the creative team (including the puppeteer, director, roboticists, and designers) perceived the six emotions and embodied the emotional states in their bodies through a series of non-verbal exercises, drawing upon techniques from charades and silent cinema. The team examined the manner in which emotions are conveyed through physical expression alone, without reliance on facial features or verbal cues. The exercises were designed to enhance the team's sensitivity to the nuances of gesture, rhythm, and posture in emotional communication, thereby establishing a shared experiential foundation for the subsequent improvisational work.

We encourage the reader to view the accompanying video material at this stage to better understand the workflow (described in later sections) in practice and how expressive motions were generated (see Supplementary).

**W2 Session 2 - Director-led Puppeteer Embodied Motion Creation:** The puppeteer, under the director's guidance, manipulated the UR3e robotic arm to improvise gestures expressing a series of basic emotions. A single object was used across all emotional contexts. Directing prompts and strategies were adjusted iteratively based on observed robot behaviour and feedback from the creative team.

This session produced a concise but diverse gesture library, recorded as time-stamped joint trajectories with metadata, forming a foundation for autonomous robotic playback. Here, autonomy refers to the robot triggering and sequencing gestures without real-time human control, based on cues, prompts, or internal logic. The library designed for reuse in *interactive performance* settings, where the robot may engage with human performers or audiences as a semi-autonomous, expressive participant within a live, dramaturgically framed environment.

Table 1 provides an overview of selected directing prompts and strategies, the corresponding robotic behaviours observed, and relevant design insights. These examples are inherently situational, shaped by the specific team involved, the dynamics of their creative collaboration, and the configuration of the robotic system. We do not claim that the exact prompts or strategies would reliably produce similar behaviours in other contexts, given the improvisational nature of the process. Rather, they are presented to illustrate the potential of integrating directing techniques and puppeteering methods to inform the co-creation of expressive robotic behaviour.

The puppeteer described the director's prompts as essential for maintaining focus and responsiveness, with clear emotional direction provided externally, he could fully concentrate on shaping the robot's movements. He particularly valued the "inner monologues" from the director, noting that the mental load required to simultaneously interpret the robot's affordances and perform expressively often left little space to generate and sustain his own imaginative context.

*Table 1.* Iterative Prompt Adjustments and Observed Effects

| Directive Prompt / Strategy | Robotic Response | Design Insight for Performative HRI |
|---|---|---|
| Iterative fear buildup *"it's gross"* → auditory "Bang!" → backstory *"ball that hurt you"* | Slow cautious approach, repeated recoil, tension release contrast retained in replay | Shows how layered cues (sensory → narrative) progressively sharpen affect |
| Antagonistic escalation "You can't have the ball" → "You're not worth it!" | Rapid side-to-side lunges, large trajectories, sustained aggressive energy | Status reversal acts as a driver of anger and a template for high-energy conflict |
| Disgust via sensory hyperbole "It smells like 1000 rotten eggs" | Abrupt full body recoil, maximal distance from object, repeated avoidance | Highlights the potency of vivid olfactory imagery for instinctive withdrawal |
| Curiosity–fear ambivalence "It's interesting but (feels) gross" | Cautious approach, side-to-side shakes, ambiguous hesitation | Captures mixed emotion and non-discrete states—useful for designing nuanced robot affect |
| Trauma-based internal conflict: The ball "broke your springs" (past trauma) - must now move it nonetheless. | Prolonged approach–retreat oscillation | Shows narrative memory as a source of sustained tension through internal contradiction |

### 4.3    Emotion as an Iterative Performance

In those improvisations, emotional nuance was not predefined but emerged via iterative refinements, each layering new directorial cues to intensify or diversify affect. For instance, fear evolved progressively from a tactile cue ("gross") to an auditory trigger ("Bang!"), culminating in a narrative backstory ("the ball that hurt you").  Disgust, initially vague under generic prompts, sharpened when anchored to sensory hyperbole ("1000 rotten eggs"), evoking reflexive withdrawal distinct from fear. The refinements were preserved in autonomous replays, particularly in fear sequences, where slow approaches followed by abrupt recoils effectively conveyed tension–release dynamics.

Complex behaviours arose particularly from conflicting and antagonistic directives. Fear prompts combined task-driven advances ("move the ball") with hesitant reactions ("it hurt last time"), generating oscillatory movements suggestive of internal struggle. Anger intensified through conflicting suggestions, affirmation ("this is the best ball") paired with denial ("you cannot have it," "you are not worth that ball"), producing a sense of exclusion that manifested as sustained aggression in rapid gestures. Scaffolded building of joy, started with free improvisation, later supported with background music ("Don't Worry, Be Happy"), stabilised spontaneous expressions of joy into rhythmic, sustained swaying motions. These strategies were not predetermined but evolved in response to the results of each iteration, shaped by ongoing assessments of the robot's performance, the director's instincts, the designers' intentions, and the dynamics between puppeteer and machine. Different teams (comprising other directors, puppeteers, and roboticists, or employing alternative robotic platforms) would likely produce distinct gesture repertoires, influenced by variations in narrative framing, embodied interpretation, and hardware affordances.

## 5    Discussion

Our practice-led approach contributes to a growing body of work integrating theatrical methodologies into human–robot interaction, proposing a structured, rehearsal-informed workflow. In contrast to prior systems using puppeteering via pre-scripted animations [15] or virtual prototyping via mixed-reality staging [17], our workflow advances HRI puppeteering by integrating three novel elements: (1) director-led improvisation using established theatre techniques, (2) real-time puppeteer–robot coupling, and (3) the potential for curated, reusable gesture libraries.

### 5.1    Dramaturgy in Emergent Robotic Expression

Hoffman's [10] conceptual suggestions are directly operationalized in our system. *Psycho-physical unity* and *mutual responsiveness* emerge as the puppeteer's kinaesthetic impulses drive robotic motion, bridging intention and actuation into a real-time feedback loop of continuous interaction. *Objectives and Obstacles* were most effective when paired. *Objectives* alone (e.g., "you want the ball") were less expressive than when coupled with *Obstacles* (e.g., "you want the ball, but you can't have it"), a technique from the Stanislavski System [21] that sharpens dramatic tension.

*Context and Given Circumstances* were primarily shaped by the director, who established narrative backstories and situational framing, such as "the ball that hurt you" and the loud "bang" noise. In contrast, the constraints and affordances of the robotic hardware and software, while not part of the fictional *Given Circumstances*, influenced expressive possibilities in a manner analogous to costume or prop limitations for an actor: they shaped how the puppeteer–robot pair could realize intentions. Thus, both dramaturgical and technical factors consistently informed improvisational choices. In this context, *improvisation* refers specifically to the rehearsal process, where it functioned as a generative method for developing expressive robotic gestures and refining interaction strategies, rather than as a live-performance response mechanism.

Perhaps the most distinctive technique emerged when the puppeteer invited the director to provide a continuous spoken stream of intention. The director articulated an externally authored "inner monologue," which the puppeteer internalised, reframed as motive, and rendered as robotic motion. This strategy was crucial for scaffolding the puppeteer's cognitive load, supporting emotional intent while enabling precise motor control, resulting in a layered enactment and distributed expressivity.

A further insight emerging from our process concerns the generative role of contradiction and conflict in shaping robotic expressivity. Long embedded in dramaturgical theory from Aristotle and Stanislavski to Pritner and Walters [21, 22], conflicting objectives and evolving circumstances serve not as disruptions but as structuring forces for transformation. Within our workflow, expressive richness often arose from tension: between intention and constraint, action and obstruction, or overlapping narrative demands. These tensions, embedded in character goals or situational dilemmas, did not inhibit expressivity, they intensified it. Yielding gestures that were affectively charged, narratively anchored, and dramaturgically meaningful.

This responsiveness to given circumstances and layered tension complements and extends prior work on emergent agency in human–robot interaction. Specifically, our approach echoes Gemeinboeck's relational-performative aesthetic, in which social behaviour unfolds through situated encounters shaped by interaction [18]. Yet, whereas her work foregrounds abstraction and open-endedness, our method reintroduces narrative scaffolds as productive constraints, anchoring improvisation in intention without closing it into fixed representation.

### 5.2    Scope and Limitations

At its core, our study explores distributed psycho-physical unity in HRI, a framework in which human intention and motive are "lent" to the machine, with theatre acting as an integrating force to weave fragmented agencies into a coherent, improvised performance. We tested this through a focused ensemble, one director, one puppeteer, two roboticists, and two designers, yielding practice-led insights into the iterative co-creation of expressive gestures. Three interdependent factors anchor the approach: kinaesthetic attunement (synchronising bodily intention with robotic motion), shared dramaturgical sensibility (aligning narrative intent across collaborators), and real-time feedback mechanisms. Emotional cues (e.g., "fear," "grief") served as internal prompts for improvisation rather than fixed labels for audiences.

Although specific outcomes will vary with team composition, hardware setup, and emotional framing, the method suggests a transferable structure for collaborative human–robot performance-making that can adapt to diverse creative and technical contexts. Future work may build on this foundation by involving more varied creative teams, integrating cross-cultural emotion frameworks, adapting the workflow to support a full creative process, and evaluating audience perception in live settings.

The robot's mechanical responsiveness, safety thresholds, and joint behaviour not only determined which gestures could be performed but also how confidently the puppeteer could execute them. These technical limitations affected expressive range and risk tolerance, especially in emotionally charged scenarios. The following section explores these challenges, comparing how various robotic platforms supported or hindered gesture expression in live contexts. This discrepancy (where puppeteers hold back from certain motions) may reveal an artistic opportunity: amplifying robotic movements beyond human comfort zones could enhance emotional expressivity.

### 5.3    Robotic Resistance and Gesture Interruption

Throughout the improvisation sessions, several gestures emerged not only through the puppeteer's intentional movement design but also in response to the robot's physical constraints and resistances. These constraints (such as torque limits, joint speed, and positional feedback) sometimes caused interruptions in gesture execution, such as premature halts, unexpected delays, or exaggerated rebounds. In particular, subtle actions like tremors or quick reversals were often dampened or distorted, producing gestures that deviated from the puppeteer's original intent.

In W1, which used the Franka Emika Panda arm in gravity-compensation mode, the puppeteer encountered frequent instances where the robot's torque or velocity thresholds were triggered unpredictably. This often resulted in abrupt joint locking that interrupted the gestural flow and required manual resetting by the roboticist. The puppeteer described these moments in anthropomorphic terms: *"When it goes into maximum, hits this point, it just stops to save itself from breaking, which is what a human would do: 'Ah! Stop, back away, I might break my arm.'" (Puppeteer, W1)*. While this framing demonstrates an intuitive understanding of the robot's protective mechanisms, it also highlights a key limitation: the absence of tactile or visual cues to help the performer anticipate such limits in real-time. This led to a subtle but persistent constraint on expressive risk-taking, particularly during delicate or emotionally charged gestures.

In contrast, the UR3e platform used in W2 was described as substantially easier to manipulate. *Freedrive* mode enabled smoother, uninterrupted joint movement, and the joint limits were less restrictive and more forgiving in live manipulation contexts, leading to a greater sense of control and expressive confidence. One standout feature was the continuous end-effector rotation, which allowed the puppeteer to maintain directional focus during complex movements, critical when simulating gaze or attention. As the puppeteer reflected, *"There's these imaginary eyes on it… and when it's not aligned, it doesn't quite 'see' the object. That middle positioning becomes really important." (Puppeteer, W2)*.

These reflections highlight how expressive manipulation is shaped by a robot's physical affordances and joint behaviour. Unpredictable constraints hinder exploration, while predictable, fluid motion enables more emotionally resonant behaviours. Designing for expressive HRI should focus on software, motion capture fidelity, and the manipulation experience itself. Performance-based robots must be responsive partners, offering freedom, tactile clarity, and transparent feedback for emotional subtlety. Future research should explore gesture-friendly modes, clear joint-limit signalling, and design heuristics that align with how humans perceive and interpret physical resistance. This invites broader reflection on the artistic affordances of human–robot motion discrepancy. What if, instead of trying to minimise or correct these divergences, designers were to embrace them? Could allowing the robot to exaggerate certain motions (through overcommitment, delay, or recoil) amplify the affective impact of gestures in emotionally charged scenes such as fear, anger, or surprise? Alternatively, could such distortions risk undermining narrative coherence or believability by making the robot appear awkward or uncontrolled?

This tension between human intention and robotic execution suggests a fertile space for performative exploration. When framed dramaturgically, motion interruptions and exaggerations may enhance expressivity by drawing attention to the robot's materiality and effort. Rather than aiming for seamless mimicry of human movement, performance robotics might benefit from leveraging the robot's distinct embodiment, acknowledging its resistance and rhythm as part of its character. These insights open up a generative area of design, where expressive nuance is not simply programmed but emerges through negotiation between human gesture, robotic capacity, and creative framing.

## 6    Conclusions

By positioning robots as puppeteered performers within a theatrical paradigm, *theatre-in-the-loop* proposes a practice-led framework for socially expressive human-robot interaction, with applications in interactive art and storytelling. Through two structured workshops, we demonstrated how director-led, puppeteer improvisation and robotic affordances can co-produce gesture libraries annotated with emotional arcs that bridge technical and artistic design. While hardware limitations occasionally constrained expressive range, our findings suggest that rehearsal-driven co-creation can yield emotionally resonant motion templates.

Future work will focus on refining these templates into cohesive performances and integrating them into live staging contexts, evaluated through audience studies. We also propose exploring generative systems that dynamically reconfigure recorded gestures while preserving their narrative intent, e.g., adapting a "joyful" gesture to varying intensities or scales. Longer-term research could investigate how robotic traits, such as mechanical precision, non-human movement logic, or modular reconfiguration, might inspire new dramaturgical vocabularies. Ultimately, this work underscores the potential of interdisciplinary collaboration between theatre and HRI to expand both the expressive capacity of robots and the creative toolkit of artists.

**Author Contribution Statement.** PP and VZHN contributed equally to the conception, development, analysis and writing of this manuscript, and share first authorship. SM is the professional puppeteer and generated the data and provided his perspectives for the manuscript. RP and RR contributed to the conceptual framing, provided the artistic scope, and participated in the workshops. AC contributed to the broader project and provided editorial feedback on the paper. AK contributed to the conception and planning of the workshops, participated in the workshops, and provided supervision and critical feedback on the manuscript.

**Acknowledgments.** This work was supported by UKRI - The Engineering and Physical Sciences Research Council (EPSRC), Centre for Doctoral Training in Horizon [grant number EP/S023305/1], the Turing AI World Leading Researcher Fellowship: Somabotics: Creatively Embodying Artificial Intelligence [grant number EP/Z534808/1], and AI UK: Creating an International Ecosystem for Responsible AI Research and Innovation (RAKE) [EP/Y009800/1]. We would also like to thank Dr Jocelyn Spence, Dr Nils Jaeger, and Dr Paul Tennent for their guidance and support in the development of the manuscript.

**Disclosure of Interests.** The authors have no competing interests to declare that are relevant to the content of this paper.

**Data access statement.** The data that support the findings of this study may be available on request from the corresponding author.

# References

1. Patel, R., Ramchurn, R.: Thingamabobas – Makers of Imaginary Worlds, https://makersofimaginaryworlds.co.uk/projects/thethingambobas/, last accessed 2023/12/20.
2. Ngo, V.Z.H., Patel, R., Ramchurn, R., Chamberlain, A., Kucukyilmaz, A.: Dancing with a Robot: An Experimental Study of Child-Robot Interaction in a Performative Art Setting. Lecture Notes in Computer Science. 15563 LNAI, 340–353 (2025). https://doi.org/10.1007/978-981-96-3525-2_29.
3. Panteris, M., Manschitz, S., Calinon, S.: Learning, generating and adapting wave gestures for expressive human-robot interaction. ACM/IEEE International Conference on Human-Robot Interaction. 386–388 (2020). https://doi.org/10.1145/3371382.3378286.
4. Zhou, A., Hadfield-Menell, D., Nagabandi, A., Dragan, A.D.: Expressive Robot Motion Timing. ACM/IEEE International Conference on Human-Robot Interaction. Part F127194, 22–31 (2017). https://doi.org/10.1145/2909824.3020221.
5. Huang, P., Hu, Y., Nechyporenko, N., Kim, D., Talbott, W., Zhang, J.: EMOTION: Expressive Motion Sequence Generation for Humanoid Robots with In-Context Learning. (2024).
6. Tang, S., Dondrup, C.: Gesture Generation from Trimodal Context for Humanoid Robots. 1, (2024). https://doi.org/10.1145/3687272.3690905.
7. Schreiter, T., Rudenko, A., Rüppel, J. V., Magnusson, M., Lilienthal, A.J.: Multimodal Interaction and Intention Communication for Industrial Robots. (2025). https://doi.org/10.1177/02783649241274794.
8. Mahadevan, K., Chien, J., Brown, N., Xu, Z., Parada, C., Xia, F., Zeng, A., Takayama, L., Sadigh, D.: Generative Expressive Robot Behaviors using Large Language Models. ACM/IEEE International Conference on Human-Robot Interaction. 482–491 (2024). https://doi.org/10.1145/3610977.3634999.
9. Katevas, K., Healey, P. G. T., & Harris, M. T. Robot Comedy Lab: experimenting with the social dynamics of live performance. *Frontiers in Psychology, 6*, 1253 (2015). https://doi.org/10.3389/fpsyg.2015.01253
10. Hoffman, G.: Hri: Four lessons from acting method. (2015). https://shorturl.at/iRgk1
11. Meisner, S., Longwell, D.: Sanford Meisner on acting. Vintage (2012).
12. Guest, A.H.: Labanotation : The System of Analyzing and Recording Movement. Routledge (2015). https://doi.org/10.4324/9780203823866.
13. Laviers, A., Maguire, C.: The BESST System: Explicating a New Component of Time in Laban/Bartenieff Movement Studies Through Work With Robots. ACM International Conference Proceeding Series. Par F180475, (2022). https://doi.org/10.1145/3537972.3538023.
14. Billard, A., Calinon, S., Dillmann, R., & Schaal, S. Survey: Robot programming by demonstration. Springer handbook of robotics, 1371-1394 (2008).
15. Hoffman, G., Kubat, R., Breazeal, C.: A hybrid control system for puppeteering a live robotic stage actor. Proceedings of the 17th IEEE International Symposium on Robot and Human Interactive Communication, RO-MAN. 354–359 (2008). https://doi.org/10.1109/ROMAN.2008.4600691.
16. Zeglin, G., Walsman, A., Herlant, L., Zheng, Z., Guo, Y., Koval, M.C., Lenzo, K., Tay, H.J., Velagapudi, P., Correll, K., Srinivasa, S.S.: HERB's Sure Thing: A rapid drama system for rehearsing and performing live robot theater. Proceedings of IEEE Workshop on Advanced Robotics and its Social Impacts, ARSO. 2015-January, 129–136 (2015). https://doi.org/10.1109/ARSO.2014.7020993.
17. Rozendaal, M.C., Vroon, J., Bleeker, M.: Enacting Human-Robot Encounters with Theater Professionals on a Mixed Reality Stage. ACM Trans Hum Robot Interact. 14, (2024). https://doi.org/10.1145/3678186.

18. Gemeinboeck, P.: The Aesthetics of Encounter: A Relational-Performative Design Approach to Human-Robot Interaction. Front Robot AI. 7, (2021). https://doi.org/10.3389/frobt.2020.577900.
19. will-d-chen/ur3e-codebase: A ros2 package to control a UR3E robot arm in joint space and task space using moveit2, can be used for any URe series arm, https://github.com/will-d-chen/ur3e-codebase, last accessed 2025/04/15.
20. Ekman, P.: Are there basic emotions? Psychol Rev. 99, 550–553 (1992). https://doi.org/10.1037/0033-295X.99.3.550.
21. Stanislavsky, K., Hapgood, E.R.: An Actor Prepares. Routledge, New York (2003).
22. Pritner, C., & Walters, S. E. *An introduction to play analysis* (2nd ed.). Waveland Press. (2017).